\documentclass{article} % For LaTeX2e
\usepackage[final]{colm2026_conference}

\usepackage{microtype}
\usepackage{hyperref}
\usepackage{url}
\usepackage{booktabs}

\usepackage{xcolor}
\usepackage{textcomp}
\definecolor{darkgreen}{RGB}{0,150,0}

\usepackage{lineno}

\usepackage{graphicx}
\usepackage{subcaption}
\usepackage{inconsolata}
\usepackage{xspace, soul}
\usepackage{threeparttable}
\usepackage{booktabs}
\usepackage{multirow}
\usepackage{amsmath}
\usepackage{pifont}
\usepackage{amssymb,enumitem}
\usepackage{comment}
\usepackage{hyperref}
\usepackage{caption}
\usepackage{wrapfig}
\usepackage{listings}
\usepackage{adjustbox}
\usepackage{marvosym}
\usepackage{makecell}
\usepackage{wrapfig}
\usepackage[ruled]{algorithm2e}
\usepackage{marvosym}
\usepackage{soul}
\usepackage{listings}

\newcommand{\method}{\mbox{CUP}\xspace}

\definecolor{darkblue}{rgb}{0, 0, 0.5}
\hypersetup{colorlinks=true, citecolor=darkblue, linkcolor=darkblue, urlcolor=darkblue}

\title{Uncertainty as a Planning Signal: \\Multi-Turn Decision Making for Goal-Oriented Conversation}

\author{Xinyi Ling$^{1}$\thanks{\,Equal contribution, \Letter\,Corresponding author.},\, Ye Liu$^{2\,*}$,\, Reza Averly$^1$, \& Xia Ning$^{1\,2\,3\,{\text{\Letter}}}$ \\
$^1$Department of Computer Science and Engineering \\
$^2$Department of Biomedical Informatics \\
$^3$Translational Data Analytics Institute \\
The Ohio State University \\
\texttt{\{ling.303, liu.12989, averly.1, ning.104\}@osu.edu} \\
}

\begin{document}

\ifcolmsubmission
\linenumbers
\fi
\maketitle
% \def\thefootnote{}\footnotetext{* Equal contribution; \Letter\,Corresponding author.}
% \begingroup
% \renewcommand{\thefootnote}{}
% \footnotetext{* Equal contribution; \Letter Corresponding author.}
% \endgroup
% \footnotetext[]{* Equal contribution; \Letter Corresponding author.}

% agent/system/framework

\begin{abstract}
Goal-oriented conversational systems require making sequential decisions under uncertainty about the user’s intent, where the algorithm must balance information acquisition and target commitment over multiple turns. Existing approaches address this challenge from different perspectives: structured methods enable multi-step planning but rely on predefined schemas, while LLM-based approaches support flexible interactions but lack long-horizon decision making, resulting in poor coordination between information acquisition and target commitment. To address this limitation, we formulate goal-oriented conversation as an uncertainty-aware sequential decision problem, where uncertainty serves as a guiding signal for multi-turn decision making. We propose a \textbf{C}onversation \textbf{U}ncertainty-aware \textbf{P}lanning framework (\method) that integrates language models with structured planning: a language model proposes feasible actions, and a planner evaluates their long-term impact on uncertainty reduction. Experiments on multiple conversational benchmarks show that \method consistently improves success rates while requiring fewer interaction turns. Further analysis demonstrates that uncertainty-aware planning contributes to more efficient information acquisition and earlier confident commitment.

\end{abstract}

%%%%%%%%%%%%%%%%%%%%%%%%%%%%%%%%%%%%%%%%%%%%%%%%%%%%%%%%
\section{Introduction}
\label{sec:introduction}
%%%%%%%%%%%%%%%%%%%%%%%%%%%%%%%%%%%%%%%%%%%%%%%%%%%%%%%%

% \begin{wrapfigure}{r}{0.475\textwidth}
%   \centering
%   \vspace{-3mm}
%   \includegraphics[width=0.425\textwidth]{figure/uncertainty.intro.pdf}
%   \caption{An illustration example of the goal-oriented conversational system. \xia{I do not think this figure is necessary}}
%   \vspace{-3mm}
%   \label{fig:intro}
% \end{wrapfigure}

Goal-oriented conversational systems aim to identify a user's underlying intent through multi-turn interactions and commit an appropriate target, such as recommending an item or completing a task~\citep{bordes2017learning, peng2024ecellm, liu2025dellma}. A central challenge lies in decision-making under uncertainty about the user's intent: the system must balance acquiring additional information and target commitment --- the decision to stop asking and commit to a specific target from candidate hypotheses. Asking more questions can reduce uncertainty but incur interaction cost, while committing too early risks incorrect outcomes~\citep{zhang2025modeling, kobalczyk2025activeatd}. Effective systems must therefore consider the long-term impact of actions on uncertainty over the user's intent.

Existing approaches address this problem from different perspectives. Traditional structured methods model dialogue as a sequential decision-making process with explicit state and action spaces, enabling multi-step planning and long-horizon optimization~\citep{bordes2017learning, wu2019transferable, wu2023goal}. However, they typically rely on predefined schemas~\citep{casanueva2020efficient} and limited domain knowledge~\citep{bisk2020experience}, which restrict their applicability in open-ended settings. More importantly, such structured formulations inherently impose rigid representations on the state and action space, limiting their ability to capture rich semantic information and adapt to diverse, free-form user inputs~\citep{bocklisch2017rasa}.

In contrast, with the rise of large language models (LLMs), LLM-based approaches have demonstrated strong performance across a wide range of conversational tasks~\citep{liu2024chatqa, yi2025survey}. They leverage rich linguistic knowledge to enable flexible and generalizable interactions, but lack explicit mechanisms for optimizing over multi-turn interaction trajectories~\citep{wei2022cot, wang2023selfconsistency, yao2023react}. Consequently, their decision-making is often driven by local or heuristic strategies, without explicit reasoning about the long-term impact of actions.

Despite these advances, traditional structured methods remain constrained by predefined schemas, while LLM-based approaches lack explicit mechanisms for multi-turn decision making, making it difficult for both to effectively coordinate information acquisition and target commitment over multiple turns. To address this limitation, we formulate goal-oriented conversation as an uncertainty-aware sequential decision problem, where an LLM is used to propose feasible actions while an uncertainty-guided planning module evaluates their long-term impact.

Specifically, we propose a \textbf{C}onversation \textbf{U}ncertainty-aware \textbf{P}lanning (\method) framework with three components: (1) a belief and uncertainty modeling module that maintains a distribution over candidate hypotheses and estimates the current uncertainty about the user's intent; (2) an uncertainty-guided planning module that evaluates proposed actions based on their long-term impact; and (3) a language-grounded execution module that realizes the selected action in natural language and updates the belief state based on the user response. 
% \xy{Unlike a simple pairing of tree search with an uncertainty score, \method treats belief uncertainty as the single principle that governs the entire decision loop—driving the search prior, the trajectory reward, and the commitment trigger alike.} 
At each turn, the agent selects actions by considering their informativeness and long-term effects, enabling more effective multi-turn decision making.

We evaluate our approach on four datasets across multiple model backbones. Experimental results show that our method consistently improves the success rate while requiring fewer interaction turns. Further analysis shows that uncertainty-aware planning leads to more efficient information acquisition and earlier confident decisions. 

Our contributions are as follows:
\begin{itemize}[leftmargin=10pt]
\item We identify uncertainty as a key signal in goal-oriented conversational decision making and investigate how it can guide multi-turn interaction.
\item We formulate goal-oriented conversation as an uncertainty-aware sequential decision problem and propose the \method framework, which leverages expected information gain for long-horizon decision making.
\item We empirically demonstrate consistent improvements in both effectiveness and efficiency across multiple datasets and backbones. 
The code is available in \href{https://github.com/ninglab/CUP}{https://github.com/ninglab/CUP}.
\end{itemize}
% 

%%%%%%%%%%%%%%%%%%%%%%%%%%%%%%%%%%%%%%%%%%%%%%%%%%%%%%%%
\section{Related Work}
\label{sec:related_work}
%%%%%%%%%%%%%%%%%%%%%%%%%%%%%%%%%%%%%%%%%%%%%%%%%%%%%%%%

% \paragraph{LLM-based Planning and Decision Making.}
% LLMs have been increasingly used in planning and decision-making and enable deliberative reasoning beyond single-step generation. 
% % Chain-of-thought (CoT)~\citep{wei2022cot} and self-consistency~\citep{wang2022self} improve decision-making by generating and aggregating intermediate thoughts.
% Previous work has explored the structured generation processes~\citep{besta2024graph},
% as well as action-interleaved interaction frameworks~\citep{yao2023react}. 
% Despite these advances, most approaches focus on improving single-step generation~\citep{wei2022cot} or rely on implicit heuristics when extended to multi-step scenarios~\citep{yao2023tot}.
% %
% More closely related are planning-oriented methods~\citep{liu2023llm+, deng2024ppdpp, loula2025syntactic}, 
% incorporate scoring signals or value estimation to guide generation. 
% However, they typically rely on heuristic decision rules without explicitly optimizing over future interaction trajectories.
% %
% In contrast, we introduce a principled planning signal to optimize
% long-term outcomes by
% balancing information acquisition and decision commitment.
% \xia{I do not think this is necessary to discuss planning and decision making, unless you have experiments comparing the related work here. }

\paragraph{Uncertainty in Active Information Acquisition.}
Active information acquisition studies how an agent should sequentially gather observations or features
to reduce its uncertainty in a downstream decision~\citep{he2016activeinformationacquisition, covert2023learning, javdani2014near}.
This problem has been studied across Bayesian experimental design~\citep{mazzaccara2024learning, choudhury2026bedllm}, sequential decision-making~\citep{kaelbling1998planning}, and active learning~\citep{settles2009active}.
Goal-oriented conversation can be viewed as a sequential decision-making problem~\citep{chopra2025misq} in which each interaction turn can serve as an
information-gathering step.
Recent works in conversational information acquisition use Monte Carlo approximation~\citep{kobalczyk2025activeatd} or probability-based uncertainty criteria~\citep{hu2024uot, zhang2025clarify} to guide decision making. Related tradeoffs between information seeking and target commitment are also studied in interactive reasoning~\citep{li2024mediq, grand2026shoot} and in LLM-based hypothesis generation with sequential Monte Carlo~\citep{piriyakulkij2024doing}.
However, most existing approaches treat uncertainty as a myopic scoring function for a single next action.
In contrast, \method incorporates uncertainty directly into tree-search priors, allowing it to guide multi-step planning rather than only greedy one-step selection.

\paragraph{Goal-Oriented Conversational Systems.}
Goal-oriented conversational systems aim to accomplish tasks through multi-turn interactions. 
Early approaches formulate conversation as a sequential decision process using slot-filling pipelines~\citep{peng2017composite}, later extended to reinforcement learning~\citep{deng2021unified, wang2025rec} and Markov decision process (MDP) frameworks for optimizing long-term task success~\citep{zhang2001spoken, williams2007partially, wen2017network, sun2018conversational, lei2020estimation}. 
Attribute-based methods further improve information acquisition by progressively refining a candidate search space~\citep{zhang2018towards, wu2019proactive, xu2021adapting}, 
while recent work explores proactive conversation strategy and modular policy planners~\citep{deng2023prompting, tang2019target, deng2024ppdpp, yu2023prompt}.
A closely related line of work builds clarification models via supervised or contrastive training on curated clarification data~\citep{chi2024clarinet,eisenstein2024dont,zhang2025modeling,chen2025learning}.
Despite modeling dialogue as a sequential process, many approaches rely on locally optimized policies and do not explicitly model long-horizon interaction trajectories, limiting coordination across turns~\citep{jannach2021survey, afchar2022explainability}. 
\method sits within this lineage as a training-free refinement that addresses this gap through structured multi-step conversational planning.

%%%%%%%%%%%%%%%%%%%%%%%%%%%%%%%%%%%%%%%%%%%%%%%%%%%%%%%%
\section{Method}
\label{sec:method}
%%%%%%%%%%%%%%%%%%%%%%%%%%%%%%%%%%%%%%%%%%%%%%%%%%%%%%%%

%===============================================
\subsection{Problem Formulation}
\label{sec:problem_formulation}
%===============================================

% \begin{wrapfigure}{r}{0.62\textwidth}
%   \centering
%   \vspace{-6mm}
%   \includegraphics[width=0.6\textwidth]{figure/uncertainty.overview.pdf}
%   \vspace{-1mm}
%   \caption{Problem Formulation}
%   \vspace{-3mm}
%   \label{fig:probelm_formulation}
% \end{wrapfigure}

\begin{wrapfigure}{r}{0.6\textwidth}
  \centering
  \vspace{-6mm}
  \includegraphics[width=0.58\textwidth]{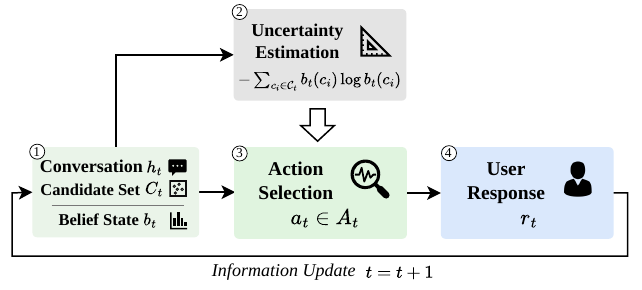}
  \vspace{-1mm}
  \caption{Problem Formulation}
  \vspace{-2mm}
  \label{fig:probelm_formulation}
\end{wrapfigure}

We model goal-oriented conversational planning as a partially observable sequential decision-making problem~\citep{williams2007partially, mrkvsic2017neural}.
Specifically, the agent maintains a belief state $b_t$ over a candidate set $C_t$ conditioned on the current conversation history $h_t$, as illustrated in Figure~\ref{fig:probelm_formulation}. At each turn $t$, the agent first estimates the uncertainty over the current belief state $b_t$, and then selects an action $a_t$ to interact with the user. With the user response $r_t$, the agent further updates the conversation history $h_{t+1}$, candidate set $C_{t+1}$ and the belief state $b_{t+1}$ for the next turn. 
This process forms an iterative decision loop, where uncertainty guides action selection, 
enabling the agent to balance information acquisition and target commitment over multiple turns.

%===============================================
\subsection{\method Overview}
\label{sec:method_overview}
\begin{figure}[t]
    \centering
    \includegraphics[width=\textwidth]{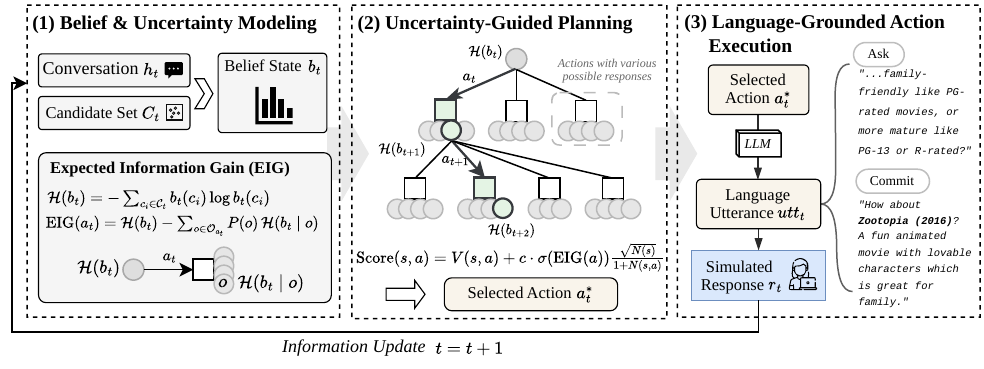}
    % \vspace{-5mm}
    \caption{The overview of \method framework.
    % \xia{make sure to refer to subfigure (1), (2) and (3) when you introduce the corresponding component}
    }
    % \vspace{-5mm}
    \label{fig:overall}
\end{figure}

%===============================================
Our proposed \textbf{C}onversation \textbf{U}ncertainty-aware \textbf{P}lanning framework (\method) consists of three components, as illustrated in Figure~\ref{fig:overall}: (1) a belief and uncertainty modeling module that maintains a belief state over candidates and estimates uncertainty; (2) an uncertainty-guided planning module that evaluates proposed actions based on their long-term impact; and (3) a language-grounded action execution module that realizes the selected action in natural language and updates the belief state based on the user response.

\subsection{Belief \& Uncertainty Modeling}
\label{sec:uncertainty}
At each turn, \method maintains a belief state $b_t$ over the candidate set $\mathcal{C}_t$, representing a probability distribution conditioned on the conversation history $h_t$.  
At initialization, the belief is computed based on the semantic similarity between each candidate $c_i \in \mathcal{C}_0$ and the initial conversation history $h_0$, which contain the user's initial query or an ongoing conversation between the user and the agent, and is never empty.
\begin{equation}
\label{eq:belief_intial}
b_0(c_i) = \frac{\exp(\mathrm{sim}(c_i, h_0))}{\sum_{c_j \in \mathcal{C}_0} \exp(\mathrm{sim}(c_j, h_0))}.
\end{equation}
Here, $\mathrm{sim}(\cdot)$ denotes cosine similarity between the embeddings of the candidate and the conversation history, assigning higher probability to candidates that are more semantically aligned with the user's intent. An ablation replacing this cosine initialization with a uniform prior is provided in Appendix~\ref{sec:ablation_uniform_init}.
% \xia{rationale of calculating in this way}
After that, we quantify the uncertainty using the entropy~\citep{seidenfeld1986entropy} of the belief state $b_t$:
\begin{equation}
\label{eq:uncertainty}
    \mathcal{H}(b_t) = - \sum_{c_i \in \mathcal{C}_t} b_t(c_i)\log b_t(c_i).
\end{equation}
Entropy measures how dispersed the belief state is over the candidate set. High entropy indicates that the belief is spread across many candidates and the agent is uncertain about the user's intent, while low entropy indicates that the belief is concentrated on a few candidates. This makes entropy a principled and widely-used gauge~\citep{namdari2019review, hu2024uot} of uncertainty for guiding conversational planning.
% \xia{explain here what entropy represents intuitively, what it indicates (high or low), why it is a good way to quantify entropy, and what entropy you would expect}

\paragraph{Commitment Trigger.}
The belief state triggers a commitment when $\frac{\mathcal{H}(b_t)}{\log |\mathcal{C}_t|} < \epsilon$ and $\max_{c_i \in C_t} b_t(c_i) \ge \theta$, indicating that the belief is sufficiently concentrated and confident. Commitment is also triggered when $|\mathcal{C}_t| \le 2$ or the maximum turn $T$ is reached. 
Otherwise, \method will proceed to the next expected information gain calculation. 

\paragraph{Expected Information Gain.}
Subsequently, we construct an action set $\mathcal{A}_t$ using an LLM (denoted as $\mathtt{LLM}$) conditioned on the current conversation history $h_t$ and candidate set $C_t$:
\begin{equation}
\mathcal{A}_t = \mathtt{LLM} (h_t, C_t),
\end{equation}
Each action $a_t \in \mathcal{A}_t$ is either (1) \textit{ask}, which queries an attribute of the candidates with several provided options; (2) \textit{commit}, which selects a candidate $c^* \in C_t$ to present to the user.
We compute the expected information gain (EIG)~\citep{lindley1956measure, ryan2003estimating} on each action $a_t \in \mathcal{A}_t$ to measure the expected reduction in uncertainty after taking action $a_t$:
\begin{equation}
\label{eq:eig}
\mathrm{EIG}(a_t, b_t) = \mathcal{H}(b_t) - \sum_{o \in \mathcal{O}_{a_t}} p(o \mid a_t)\,\mathcal{H}(b_t \mid o),
\end{equation}
where $\mathcal{O}_{a_t}$ denotes the set of possible observation to action $a_t$, and $\mathcal{H}(b_t \mid o)$ is the posterior entropy after observing $o \in \mathcal{O}_{a_t}$. 
The possible observation $\mathcal{O}_{a_t}$ are directly derived from the action $a_t$: for an \emph{ask} action, they correspond to the provided options; for a \emph{commit} action, they correspond to user acceptance or rejection.
EIG scores are then converted into a probability distribution over actions, serving as a prior for planning:
\begin{equation}
\label{eq:eig_prior}
P(a_t, b_t) = \frac{\exp(\mathrm{EIG}(a_t, b_t))}{\sum_{a'_t \in \mathcal{A}_t} \exp(\mathrm{EIG}(a'_t, b_t))}.
\end{equation}
% \xia{here, $b_t$ is not aggregated out -- it appears on the right side of the equation without aggregation, so it must 
% be also appearing on the left side of the equation, $P(a_t, b_t)$}
A higher $P(a_t, b_t)$ indicates that the action $a_t$ is expected to reduce uncertainty more given the current belief state,
% \xia{how is it related to $b_t$}
and is therefore prioritized during planning.

%===============================================
\subsection{Uncertainty-Guided Planning}
\label{sec:method_planning}
%===============================================
Given the current belief state $b_t$, \method selects actions via lookahead planning over future interaction trajectories. We instantiate the planner using Monte Carlo Tree Search (MCTS)~\citep{browne2012survey, chopra2025misq}, where uncertainty is used as a signal to guide multi-turn decision making.

\paragraph{MCTS Planning.}
Starting from $b_t$, MCTS builds a search tree where each node represents a belief state 
and each edge represents an action outgoing from that belief state node, simulating future trajectories. 
The rollouts inside MCTS use a lightweight deterministic surrogate simulator that resolves a hypothetical user response by attribute lookup over the current candidate set, without any LLM calls, following the surrogate-rollout design of~\citet{chopra2025misq}. This surrogate is deliberately separate from the LLM-based user simulator used for evaluation (Section~\ref{sec:exp_setup}): the planner never queries the evaluation-time simulator, which keeps planning decoupled from evaluation and per-turn LLM cost fixed as the search budget grows.
Each action $a_t \in \mathcal{A}_t$ is evaluated using the following adapted scoring function~\citep{silver2017mastering}:
\begin{equation}
\label{eq:score}
\mathrm{Score}(a_t, b_t) = \mathcal{V}(a_t, b_t) + c \cdot P(a_t, b_t)\frac{\sqrt{N(b_t)}}{1 + N(b_t,a_t)},
\end{equation}
The first term $\mathcal{V}(b_t,a_t)$ is the estimated action value, and exploits currently estimated action values. While the second term encourages exploration of less-visited actions weighted by their EIG-based prior. 
$P(a_t, b_t)$ is the EIG-based prior from Eq.~\ref{eq:eig_prior}, $N(b_t)$ denotes the visit counts of the node corresponding to belief state $b_t$, and $N(b_t,a_t)$ denotes the edge counts corresponding to action $a_t$ taken from $b_t$. 
% In this equation, the first term $\mathcal{V}(b_t,a_t)$ exploits currently estimated action values, While the second term $c \cdot P(a_t)\frac{\sqrt{N(b_t)}}{1 + N(b_t,a_t)}$ encourages exploration of less-visited actions weighted by their EIG-based prior.
% \ye{contributing to the search toward informative actions rather than exploring uniformly at random.}
%
The action value is defined as the expected cumulative reward over simulated trajectories:
\begin{equation}
\mathcal{V}(b_t,a_t) = \mathbb{E}\left[\sum_{t'=t}^{T} \gamma^{t'-t}\, R(a_{t'}, b_{t'})\right],
\end{equation}
where $\gamma \in (0,1]$ is a discount factor that down-weights rewards from future turns
and the reward $R$ balances task success, interaction efficiency, and uncertainty reduction:
\begin{equation}
\label{eq:rab}
R(a_t, b_t) = \mathbf{1}[\text{success}] - \lambda + \alpha \cdot \mathrm{EIG}(a_t, b_t) - \beta \cdot \mathbf{1}[\text{failure}].
\end{equation}
where $\mathbf{1}[\text{success}]$ and $\mathbf{1}[\text{failure}]$ indicate whether the commitment is accepted or rejected by the user, respectively, which are calculated only when $a_t$ is a \emph{commit} action.
$\lambda > 0$ is a turn-level penalty factor that accumulates over turns and penalizes longer conversations. $\alpha > 0$ weights the uncertainty reduction signal (EIG), and $\beta > 0$ penalizes incorrect commitments.

\paragraph{Action Selection.}
After planning, the best action $a^*_t = \arg\max_{a_t} \mathcal{V}(b_t,a_t)$ is selected for execution and then passed to the execution module (Section ~\ref{sec:execution}) to produce the corresponding utterance.

%===============================================
\subsection{Language-Grounded Action Execution}
\label{sec:execution}
%===============================================
The selected action $a^*_t$ is realized as a natural language utterance $utt_t$ by the $\mathtt{LLM}$:
\begin{equation}
utt_t = \mathtt{LLM}(a^*_t, h_t, C_t, b_t),
\end{equation}
where $h_t$ denotes the conversation history, $C_t$ is the current candidate set, and $b_t$ is the corresponding belief state. 
More specifically, if $a^*_t$ is an \textit{ask} action, the $\mathtt{LLM}$ will be employed to generate a question about a specific attribute, accompanied by a set of options, where $utt_t \leftarrow \{Question, Options\}$.
The options partition the candidate set, so that each possible user response eliminates a subset of inconsistent candidates and reduces uncertainty.
If $a^*_t$ is a \textit{commit} action, \method selects the most probable candidate $c^* = \arg\max_{c_i \in \mathcal{C}_t} b_t(c_i)$. When $\max_{c_i} b_t(c_i) \geq \theta$, the $\mathtt{LLM}$ produces a direct commitment. When the \emph{commit} action is selected by the planning module, or $|\mathcal{C}_t| \le 2$, or the maximum turn $T$ is reached, 
% \xia{it seems this paragraph is under the condition that $a*$ is a commit action, that means $\max_{c_i} b_t(c_i) \geq \theta$ is always true. Make it crystal clear what ``otherwise" indicates exactly} 
$\mathtt{LLM}$ is further used to perform semantic reasoning and generate a refined commitment by leveraging the conversation history $h_t$ and current candidate set $C_t$. Here, the $\mathtt{LLM}$ will identify the candidate that best matches the user's expressed preferences as $c^*$, and generate $utt_t$ committing $c^*$ to the user, denoted as $utt_t \leftarrow \{c^*\}$.

\paragraph{State Update.}
The utterance $utt_t$ is then presented to the user simulator, which returns a response $r_t$.  
% \xia{here $o$ is defined first time}
The conversation history is updated as $h_{t+1} = h_t \cup \{utt_t, r_t\}$, and the candidate set $\mathcal{C}_{t+1} \subseteq \mathcal{C}_t$ is pruned by removing candidates that contradict $r_t$. The belief state is updated as:
\begin{equation}
\label{eq:belief_update}
b_{t+1}(c_i) \propto
\begin{cases}
b_t(c_i) \cdot \mathrm{sim}(c_i, h_{t+1})^{\delta}, & c_i \in \mathcal{C}_{t+1},\\
0, & \text{otherwise,}
\end{cases}
\end{equation}
where $\delta \geq 0$ controls the influence of the updated dialogue history.
Here, $\mathrm{sim}(\cdot)$ measures the semantic similarity between $c_i$ and $h_{t+1}$. 
The update follows a Bayesian multiplicative logic, retaining the prior information in $b_t$ while reweighting each candidate with new information. Candidates contradicting the user response are filtered out. After applying the update, $b_{t+1}$ is renormalized.
The updated information seeds the next turn, and the interaction proceeds until a commitment is accepted by the user or the maximum turn budget is reached.

%%%%%%%%%%%%%%%%%%%%%%%%%%%%%%%%%%%%%%%%%%%%%%%%%%%%%%%%
\section{Experiments}
\label{sec:experiments}
%%%%%%%%%%%%%%%%%%%%%%%%%%%%%%%%%%%%%%%%%%%%%%%%%%%%%%%%

%===============================================
\subsection{Experimental Setup}
\label{sec:exp_setup}
%===============================================
\paragraph{Datasets.}
\begin{wraptable}{r}{0.5\textwidth}
\vspace{-11pt}
\centering
\caption{Dataset statistics. \#Conv. is the number of conversations. \#Attr. is the number of attributes associated with candidates.} 
\label{tbl:dataset}
\footnotesize
\begin{threeparttable}
\begin{tabular}{
  @{\hspace{4pt}}l@{\hspace{2pt}}
  @{\hspace{2pt}}r@{\hspace{2pt}}
  @{\hspace{2pt}}r@{\hspace{2pt}}
  @{\hspace{2pt}}r@{\hspace{2pt}}
  @{\hspace{2pt}}r@{\hspace{4pt}}
}
\toprule
Dataset & \#Conv. & \#Attr. & \makecell[r]{Avg. Words\\per Conv.} & \makecell[r]{Candidate\\Pool Size}  \\
\midrule
Beauty   & 772 & 8 & 481.2 & 3,392  \\
Fashion  & 826 & 8 & 403.1 & 4,034  \\
Home     & 372 & 9 & 199.3 & 1,180  \\
Inspired & 98  & 6 & 279.3 & 17,731  \\
\bottomrule
\end{tabular}
\end{threeparttable}
% \vspace{-6pt}
\end{wraptable}

We evaluate our method on four evaluation domains across two source datasets: Inspired~\citep{hayati2020inspired}, a movie-recommendation benchmark widely used in recent goal-oriented conversation research~\citep{deng2024ppdpp, zhang2025modeling}, and Beauty, Fashion, and Home, three e-commerce domains drawn from the LaViC benchmark~\citep{jeon2025adapting}. Each dataset consists of multi-turn dialogues paired with a ground-truth target.
Following prior work~\citep{hu2024uot, chopra2025misq}, we construct, for each conversation, a candidate set with a ground-truth target and 299 distractor candidates retrieved by SBERT~\citep{reimers2019sbert} from the full dataset candidate pool. 
More statistics are illustrated in Table~\ref{tbl:dataset}.
For all methods, we adopt the limited-turn protocol from prior goal-oriented conversation benchmarks~\citep{hu2024uot, chopra2025misq}: every method must commit to a final target by turn $T=5$ (or earlier if it chooses), and a method that reaches the turn limit without committing is forced to commit to its current best candidate, with an incorrect commitment counted as a failure.

\paragraph{Evaluation Metrics.}
We adopt two primary evaluation metrics~\citep{hu2024uot, chopra2025misq} that reflect the core objectives of conversational decision making, including
% \begin{itemize}[leftmargin=20pt]
    \textbf{(1) Success Rate (SR)} measures whether the system successfully commits the ground-truth target within a limited number of turns, capturing the overall accuracy.
    \textbf{(2) Average Turns (avgT)} measures the number of interaction turns in the conversation, reflecting the efficiency of the method.
% \end{itemize}

\paragraph{Baselines.}
We compare our method against a comprehensive set of baselines 
spanning retrieval-based, LLM-based, and planning-based approaches. 
We also report additional comparisons against two training-free planning baselines in Appendix~\ref{sec:additional_baselines}.
Specifically, here we include: 
\begin{itemize}[leftmargin=20pt]
\item \textbf{Retrieval-based methods}: 
\textit{SBERT}~\citep{reimers2019sbert} is a no-planning retrieval baseline. At each turn, it embeds the accumulated conversation history and ranks all candidates by cosine similarity, committing to the top-ranked candidate.
\item \textbf{Heuristic LLM-based methods}: 
\textit{Direct Prompting}, directly generates responses by an LLM without explicit reasoning or multi-turn decision making. 
\textit{Chain-of-Thought (CoT)}~\citep{wei2022cot}, improves local reasoning via chain-of-thought prompting but does not model long-horizon interaction. 
\textit{LLM planning}, performs heuristic multi-step reasoning without explicit uncertainty modeling.
\item \textbf{Uncertainty-aware planning methods}: 
\textit{UoT}~\citep{hu2024uot}, uses uncertainty over decision tree to guide question selection. 
\textit{ATD}~\citep{kobalczyk2025activeatd}, select clarification questions to reduce task ambiguity by sample-based approximation. 
\textit{BED-LLM}~\citep{choudhury2026bedllm}, applies logit-based expected information gain for informative action selection. 
\textit{MISQ-HF}~\citep{chopra2025misq}, combines uncertainty with cluster-based multi-step conversational planning.
\end{itemize}

\paragraph{Implementation Details.}
We use Qwen3-4B~\citep{yang2025qwen3}, Mistral-7B-v0.3~\citep{mistral7b-instruct-v03}, and Llama-3.1-8B~\citep{grattafiori2024llama} as backbone models. The user simulator is an instruction-prompted Llama-3.2-3B~\citep{grattafiori2024llama} conditioned on the target item, following the LLM-user-simulator protocol of prior work~\citep{sekulic2024reliable, abbasiantaeb2024let}. It answers the agent's questions and accepts or rejects commitments in natural language. A name-based leakage filter further prevents the simulator from revealing the target. Full prompt templates are provided in Appendix~\ref{sec:prompts_appendix}.
In our \method framework, 
for planning, we adopt a Monte Carlo Tree Search with search budget $K=50$ and exploration constant $c=1.4$. The discount factor is set to $\gamma=0.99$, the information gain bonus $\alpha = 0.2$, the turn penalty $\lambda = 0.1$, and the failure penalty $\beta = 0.5$. The entropy-based commitment thresholds of normalized entropy ratio and maximum belief probability are set to $\epsilon=0.5$ and $\theta=0.8$, respectively.
Belief updates use Bayesian multiplicative updates with $\delta = 1.0$. Embeddings are computed using SBERT~\citep{reimers2019sbert}. All experiments are conducted with a maximum turn limit of $T=5$. 
A hyperparameter sweep is provided in Appendix~\ref{sec:sensitivity}, and computational cost on search budget is provided in Appendix~\ref{sec:efficiency}.

%===============================================
\subsection{Main Results}
\label{sec:exp_main}
%===============================================

\begin{table*}[t]
  \centering
  \caption{Overall Performance}
  % \vspace{-4pt}
      \label{tbl:overall}

  \footnotesize
  \begin{threeparttable}
      \begin{tabular}{
        @{\hspace{4pt}}l@{\hspace{0pt}}
        @{\hspace{-5pt}}r@{\hspace{2pt}}
        @{\hspace{2pt}}r@{\hspace{0pt}}
        @{\hspace{0pt}}r@{\hspace{2pt}}
        @{\hspace{2pt}}r@{\hspace{2pt}}
        @{\hspace{2pt}}r@{\hspace{0pt}}
        @{\hspace{0pt}}r@{\hspace{2pt}}
        @{\hspace{2pt}}r@{\hspace{2pt}}
        @{\hspace{2pt}}r@{\hspace{0pt}}
        @{\hspace{0pt}}r@{\hspace{2pt}}
        @{\hspace{2pt}}r@{\hspace{2pt}}
        @{\hspace{2pt}}r@{\hspace{0pt}}
        @{\hspace{0pt}}r@{\hspace{4pt}}
      }
      \toprule
      \multirow{2}{*}{Model} & \multirow{2}{*}{Method} & 
      \multicolumn{2}{c}{Beauty} && 
      \multicolumn{2}{c}{Fashion} && 
      \multicolumn{2}{c}{Home} && \multicolumn{2}{c}{Inspired} \\
      \cmidrule{3-4} \cmidrule{6-7} \cmidrule{9-10} \cmidrule{12-13}
      && \multicolumn{1}{l}{SR(\%)} & \multicolumn{1}{l}{avgT} & &\multicolumn{1}{l}{SR(\%)} & \multicolumn{1}{l}{avgT} && \multicolumn{1}{l}{SR(\%)} & \multicolumn{1}{l}{avgT} && \multicolumn{1}{l}{SR(\%)} & \multicolumn{1}{l}{avgT} \\
      
      \midrule
      \multicolumn{2}{c}{SBERT} & 11.14 & 4.82 && 15.86 & 4.79 && 18.82 & 4.73 && 11.22 & 4.85 \\
      
      \midrule
      
      \multirow{8}{*}{\rotatebox{90}{\centering Qwen3-4B}} 
      & Direct Prompting    & 37.05 & 4.18 && 37.65 & 4.13 && 41.92 & 4.05 && 32.65 & 4.22 \\
      & CoT                 & 38.47 & 4.11 && 39.23 & 4.05 && 48.39 & 3.88 && 45.92 & 4.01 \\
      & LLM-planning        & 44.30 & 4.72 && 41.53 & 4.70 && 51.34 & 4.65 && 48.98 & 4.73 \\
      & UoT                 & 51.68 & 4.67 && 49.52 & 4.80 && 59.68 & 4.69 && 59.18 & 4.70 \\
      & ATD                 & 48.96 & 4.70 && 48.43 & 4.62 && 53.76 & 4.58 && 58.16 & 4.68 \\
      & BED-LLM             & 55.57 & 4.49 && 51.82 & 4.38 && 65.05 & 4.31 && 65.31 & 4.41 \\
      & MISQ-HF             & 63.21 & 4.33 && 56.41 & 4.20 && 69.89 & 4.10 && 67.35 & 4.30 \\
      & \method (ours)      & \textbf{82.12} & \textbf{4.10} && \textbf{83.41} & \textbf{4.16} && \textbf{84.14} & \textbf{3.67} && \textbf{89.80} & \textbf{3.93} \\

      \midrule
      \multirow{8}{*}{\rotatebox{90}{\centering Mistral-7B-v0.3}} 
      & Direct Prompting    & 32.25 & 4.20 && 33.90 & 4.18 && 39.52 & 4.11 && 27.55 & 4.23 \\
      & CoT                 & 35.42 & 4.17 && 38.14 & 4.01 && 45.98 & 3.93 && 36.73 & 4.09 \\
      & LLM-planning        & 43.01 & 4.81 && 41.65 & 4.69 && 52.69 & 4.62 && 48.98 & 4.70 \\
      & UoT                 & 53.11 & 4.73 && 50.85 & 4.77 && 59.13 & 4.71 && 61.22 & 4.66 \\
      & ATD                 & 49.22 & 4.69 && 48.67 & 4.62 && 54.30 & 4.50 && 57.14 & 4.63 \\
      & BED-LLM             & 61.53 & 4.62 && 50.24 & 4.45 && 63.44 & 4.34 && 66.33 & 4.58 \\
      & MISQ-HF             & 64.90 & 4.30 && 59.32 & 4.16 && 71.77 & 4.06 && 74.49 & 4.28 \\
      & \method (ours)      & \textbf{82.90} & \textbf{4.06} && \textbf{83.77} & \textbf{4.11} && \textbf{83.87} & \textbf{3.62} && \textbf{91.84} & \textbf{4.01} \\

      \midrule
      \multirow{8}{*}{\rotatebox{90}{\centering Llama-3.1-8B}} 
      & Direct Prompting    & 33.94 & 4.22 && 35.23 & 4.21 && 38.17 & 4.15 && 28.57 & 4.16 \\
      & CoT                 & 37.82 & 4.10 && 38.62 & 4.13 && 47.31 & 3.94 && 40.82 & 3.99 \\
      & LLM-planning        & 44.43 & 4.75 && 40.68 & 4.73 && 49.19 & 4.70 && 50.00 & 4.77 \\
      & UoT                 & 57.90 & 4.69 && 52.06 & 4.73 && 60.48 & 4.61 && 62.23 & 4.71 \\
      & ATD                 & 49.09 & 4.66 && 52.30 & 4.60 && 56.99 & 4.56 && 58.16 & 4.64 \\
      & BED-LLM             & 68.78 & 4.38 && 59.32 & 4.40 && 69.35 & 4.35 && 71.43 & 4.39 \\
      & MISQ-HF             & 69.95 & 4.31 && 63.44 & 4.18 && 75.00 & 4.12 && 77.55 & 4.32 \\
      & \method (ours)             & \textbf{81.99} & \textbf{4.06} && \textbf{82.20} & \textbf{4.01} && \textbf{89.78} & \textbf{3.75} && \textbf{92.86} & \textbf{3.86} \\
      
      \bottomrule
      \end{tabular}
  \end{threeparttable}
  % \vspace{-12pt}
\end{table*}

Table~\ref{tbl:overall} presents the overall performance, where \method consistently achieves the best results across datasets and model backbones. We highlight three key observations from the results.

\textbf{Planning with uncertainty leads to more effective multi-turn decision making.}
Methods that treat each turn independently (e.g., Direct Prompting, CoT) perform poorly, as they lack mechanisms to anticipate future interactions, resulting in myopic decisions. While LLM-based planning introduces heuristic multi-step simulation, it lacks a principled signal and often produces inefficient interaction (e.g., on Beauty with Llama-3.1-8B, LLM-planning 4.75 vs. CoT 4.10). 
Uncertainty-aware methods (e.g., ATD, UoT, BED-LLM, MISQ-HF) improve performance by prioritizing actions that reduce ambiguity, but typically rely on one-step or heuristic signals. In contrast, \method integrates uncertainty into planning as a global decision signal, enabling evaluation of actions based on their long-term impact. This leads to consistently higher success rates, especially on challenging datasets such as Inspired (Qwen3-4B: 58.16\% (ATD) → 67.35\% (MISQ-HF) → 89.80\% (\method)).

\textbf{\method's gains vary across domains.}
Across all settings, \method consistently outperforms no-planning baselines (e.g., Direct Prompting), but the magnitude of improvement differs by domain. For example, gains over Direct Prompting are largest on Inspired (Llama-3.1-8B, +64.29\%), while Fashion and Home exhibit smaller yet stable improvements (around 50\%). This difference reflects the varying difficulty of multi-turn information acquisition: Inspired involves high ambiguity and semantic overlap among candidates, making it difficult to identify the target from context alone, 
consistent with its weaker performance of SBERT. These results indicate that \method is particularly beneficial in scenarios that require sustained information acquisition over multiple turns.

\textbf{Planning reduces sensitivity to backbone models.}
The performance gap between Qwen3-4B and Llama-3.1-8B is relatively small under \method (e.g., Beauty 82.12\% vs. 81.99\%), but more pronounced for LLM-based methods (Direct Prompting: 37.05\% vs. 33.94\%). This suggests that our designed planning mechanism reduces reliance on model capacity by guiding action selection. The consistent performance across backbones further indicates that \method improves the decision process itself rather than depending on stronger LLMs.

%===============================================
\subsection{Ablation Study}
\label{sec:ablation}
%===============================================

% \captionsetup{skip=5pt}
\begin{figure}[h]
    \centering
    \includegraphics[width=0.9\linewidth]{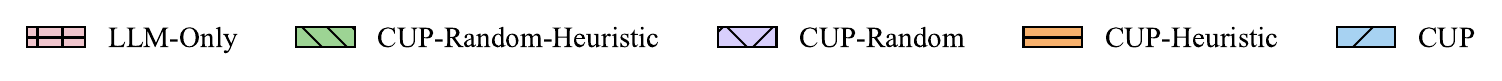}
    \\
    \begin{subfigure}[t]{0.24\textwidth}
        \centering
        \includegraphics[width=\linewidth]{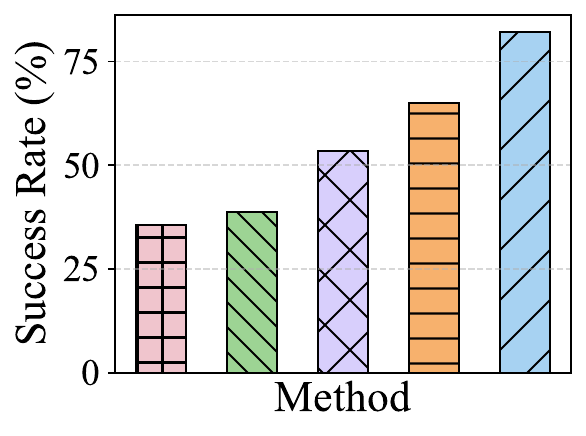}
        \caption{Beauty}
    \end{subfigure}
    \hfill
    \begin{subfigure}[t]{0.24\textwidth}
        \centering
        \includegraphics[width=\linewidth]{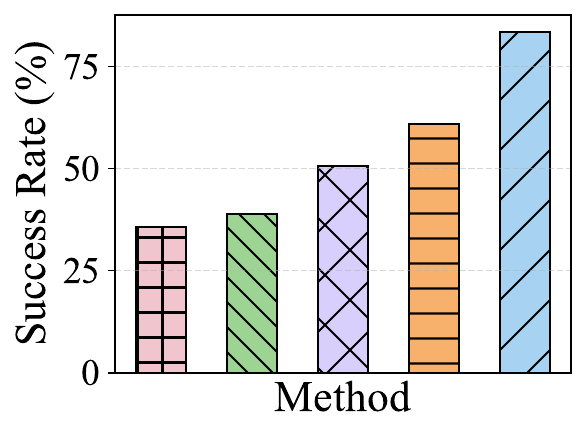}
        \caption{Fashion}
    \end{subfigure}
    \hfill
    \begin{subfigure}[t]{0.24\textwidth}
        \centering
        \includegraphics[width=\linewidth]{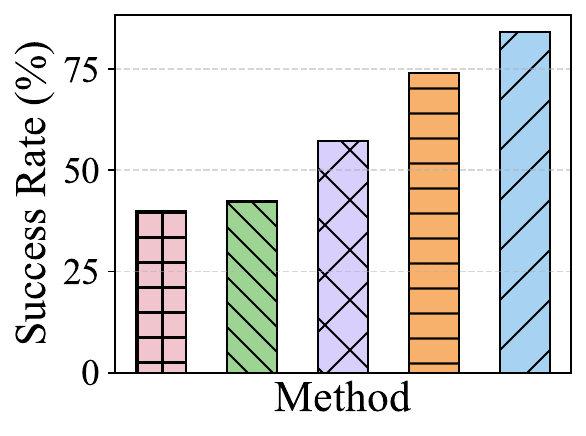}
        \caption{Home}
    \end{subfigure}
    \hfill
    \begin{subfigure}[t]{0.24\textwidth}
        \centering
        \includegraphics[width=\linewidth]{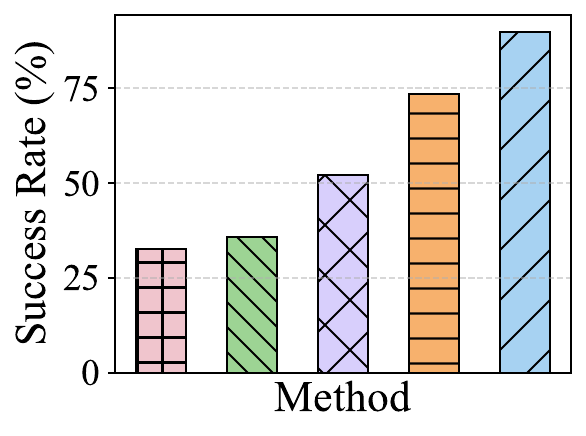}
        \caption{Inspired}
    \end{subfigure}
    % \vspace{-2mm}
    \caption{Ablation results on four datasets.}
    % \vspace{-5mm}
    \label{fig:ablation}
\end{figure}
% \vspace{-5pt}

\paragraph{Component ablation.} 
To understand the contribution of each component, we conduct a factorized ablation study that decomposes \method along two dimensions: (1) action selection (planning) and (2) commitment realization (execution). All variants share the same interaction budget and backbone (Qwen3-4B). Specifically, we consider the following configurations:
% \vspace{-3pt}
\begin{itemize}[leftmargin=20pt]
    \item \textbf{\method-Heuristic}: retains the planning module but replaces LLM-based commitment with a heuristic top-1 candidate commitment.
    \item \textbf{\method-Random}: removes structured planning by selecting actions uniformly at random, while retaining LLM-based commitment.
    \item \textbf{\method-Random-Heuristic}: removes both planning and LLM-based commitment, serving as a lower bound reference.
    \item \textbf{LLM-only}: directly generates commitment decisions using the LLM without explicit planning or structured action selection.
\end{itemize}

Figure~\ref{fig:ablation} presents the success rate across four datasets. The full \method model consistently achieves the best performance, indicating that uncertainty-aware planning and language-grounded commitment play complementary roles. Planning is the dominant factor, as variants with structured planning (\method, \method-Heuristic) consistently outperform those without it (\method-Random, \method-Random-Heuristic), while LLM-based commitment provides additional but smaller gains. In contrast, the variant LLM-only performs poorly, highlighting that local decision strategies are insufficient for effective multi-turn interaction.

\paragraph{Multi-step search vs. uncertainty signal.}
To further isolate the contributions of multi-step MCTS planning and the
EIG uncertainty signal—the two forces behind \method's planner—we introduce
two additional variants. \method-NoEIG keeps the full multi-step MCTS
planner but removes the EIG signal, i.e., both the EIG-based prior
(Eq.~\ref{eq:eig_prior}) and the reward bonus ($\alpha{=}0$ in
Eq.~\ref{eq:rab}), so the planner performs multi-step search over
LLM-proposed actions without an explicit uncertainty signal. \method-1Step
keeps the EIG signal but removes multi-step planning, replacing MCTS with a
greedy one-step $\arg\max$ over EIG. We compare these two variants against
\method-Random, which removes both and serves as the lower-bound reference,
and full \method.

\begin{wraptable}{r}{0.525\textwidth}
\centering
\vspace{-12pt}
\caption{Multi-step search vs. uncertainty signal.}
\vspace{-5pt}
\label{tbl:ingredient_ablation}
\footnotesize
\begin{threeparttable}
\begin{tabular}{
  @{\hspace{4pt}}l@{\hspace{2pt}}
  @{\hspace{2pt}}r@{\hspace{2pt}}
  @{\hspace{2pt}}r@{\hspace{2pt}}
  @{\hspace{2pt}}r@{\hspace{2pt}}
  @{\hspace{2pt}}r@{\hspace{4pt}}
}
\toprule
Variant & Beauty & Fashion & Home & Inspired \\
\midrule
\method-Random  & 53.4 & 50.6 & 57.3 & 52.0 \\
\method-1Step & 65.2 & 65.6 & 70.4 & 76.5 \\
\method-NoEIG & 72.7 & 74.5 & 75.0 & 81.6 \\
\method & \textbf{82.1} & \textbf{83.4} & \textbf{84.1} & \textbf{89.8} \\
\bottomrule
\vspace{-5pt}
\end{tabular}
\end{threeparttable}
\end{wraptable}
Table~\ref{tbl:ingredient_ablation} shows that the two are complementary:
starting from \method-Random ($\sim$50\%), adding either \method-1Step or
\method-NoEIG recovers a substantial fraction of the gap to full \method,
yet neither alone matches it. Multi-step planning provides the substrate and
the EIG signal provides the direction; both are needed to reach the top
operating point.

% \paragraph{Element ablation within the planner.}
% To further isolate two key elements of \mbox{\method}—multi-step MCTS
% planning and the EIG uncertainty signal—we introduce two additional
% variants. \method-NoEIG keeps the full multi-step MCTS planner but removes
% the EIG signal, i.e., both the EIG-based prior (Eq.~\ref{eq:eig_prior}) and
% the reward bonus ($\alpha{=}0$ in Eq.~\ref{eq:rab}), so the planner performs multi-step search
% over LLM-proposed actions without an explicit uncertainty signal.
% \method-1Step keeps the EIG signal but removes multi-step planning, replacing
% MCTS with a greedy one-step $\arg\max$ over EIG. We compare these two
% variants against \method-Random, which removes both elements and serves as
% the lower-bound reference, and full \method.

% \input{table/ingredient_ablation}
% Table~\ref{tbl:ingredient_ablation} shows that the two elements are
% complementary: starting from \method-Random ($\sim$50\%), adding either
% \method-1Step or \method-NoEIG recovers a substantial fraction of the gap to
% full \method, yet neither alone matches it. Multi-step planning provides the
% substrate and the EIG signal provides the direction; both are needed to reach
% the top operating point.

%===============================================
\subsection{Dominance of Target Candidate}
\label{sec:exp_dominance}
%===============================================

\begin{figure}[h]
% \vspace{-3mm}
    \begin{subfigure}[t]{0.24\textwidth}
        \centering
        \includegraphics[width=\linewidth]{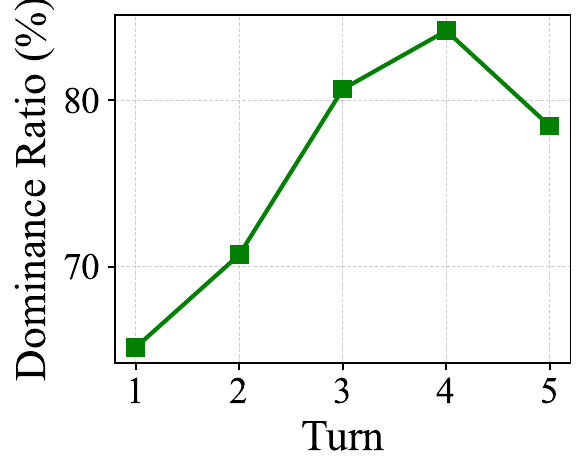}
        \caption{Beauty}
    \end{subfigure}
    \hfill
    \begin{subfigure}[t]{0.24\textwidth}
        \centering
        \includegraphics[width=\linewidth]{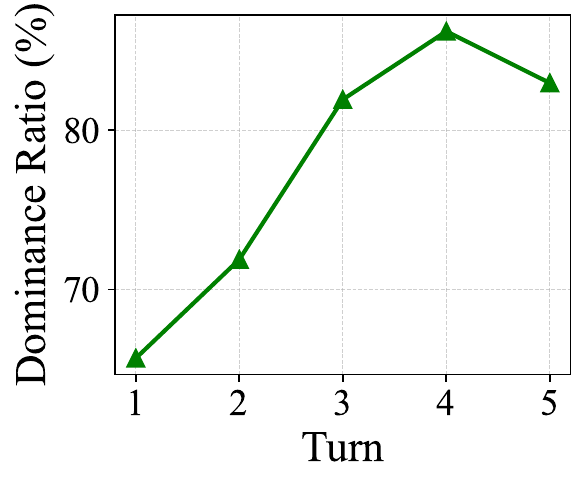}
        \caption{Fashion}
    \end{subfigure}
    \hfill
    \begin{subfigure}[t]{0.24\textwidth}
        \centering
        \includegraphics[width=\linewidth]{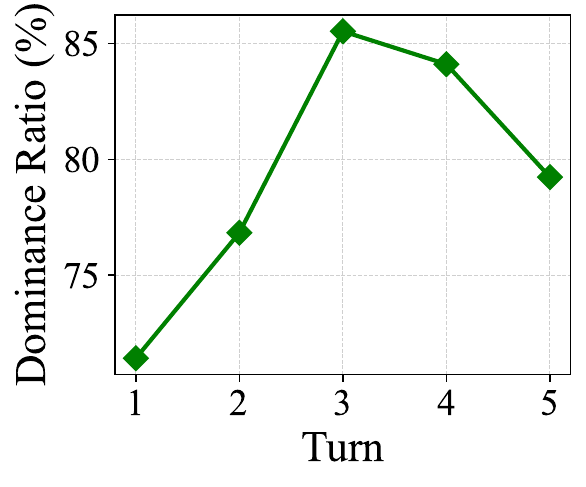}
        \caption{Home}
    \end{subfigure}
    \hfill
    \begin{subfigure}[t]{0.24\textwidth}
        \centering
        \includegraphics[width=\linewidth]{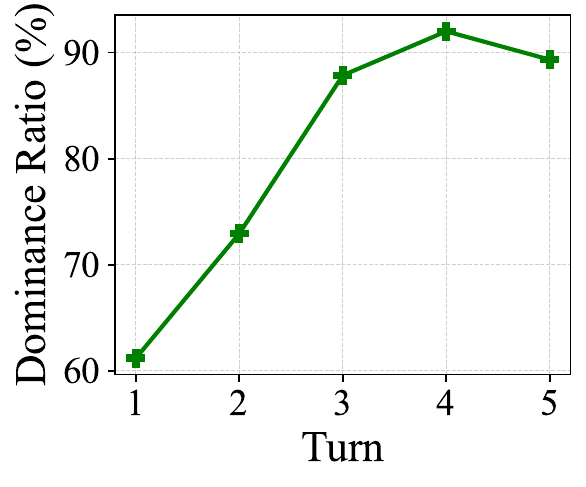}
        \caption{Inspired}
    \end{subfigure}
    % \vspace{-2pt}
    \caption{Dominance ratio of the target candidate.}
    % \vspace{-1mm}
    \label{fig:dominance}
\end{figure}

To understand how \method progressively narrows the candidate search space, we analyze the dominance of the target candidate across interaction turns. Specifically, we define the dominance ratio as:
\begin{equation}
    \text{DominanceRatio}_t = 1 - \frac{\text{Rank}_t(c^*) - 1}{|C_t|},
\end{equation}
where $\text{Rank}_t(\cdot)$ denotes the position of a candidate in the ranking induced by the belief state at turn $t$, with 1 being the highest rank. $|C_t|$ represents the size of the current candidate set. This metric measures how prominent the target candidate $c^*$ is within the remaining candidate set $C_t$, where a value of 1.0 indicates that the target is ranked first.

Figure~\ref{fig:dominance} shows that the dominance ratio increases steadily during early turns (turn 1-3) across all datasets, indicating that the target rapidly rises in rank as informative interactions are performed. This trend reflects effective candidate discrimination, demonstrating the advantage of our \method framework. A slight decline in later turns (turn 4-5) is mainly due to early commitment on confident cases, leaving more ambiguous instances in subsequent turns. 
% We also analyzed the uncertainty reduction over interaction turns, which can be found in Appendix~\ref{sec:exp_uncertainty_reduction}.

%===============================================
\section{Uncertainty Reduction Analysis}
\label{sec:exp_uncertainty_reduction}
%===============================================

\begin{figure}[h]
    \begin{subfigure}[t]{0.24\textwidth}
        \centering
        \includegraphics[width=\linewidth]{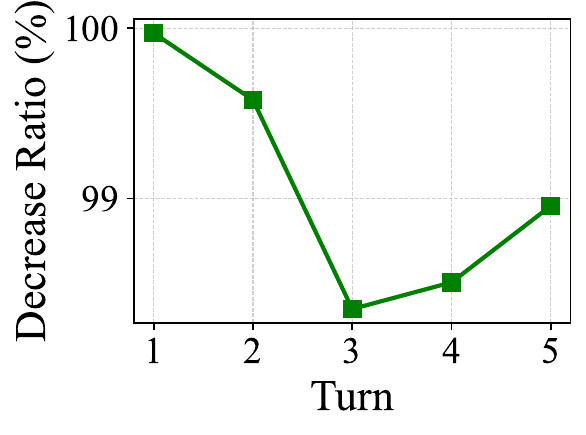}
        \caption{Beauty}
    \end{subfigure}
    \hfill
    \begin{subfigure}[t]{0.24\textwidth}
        \centering
        \includegraphics[width=\linewidth]{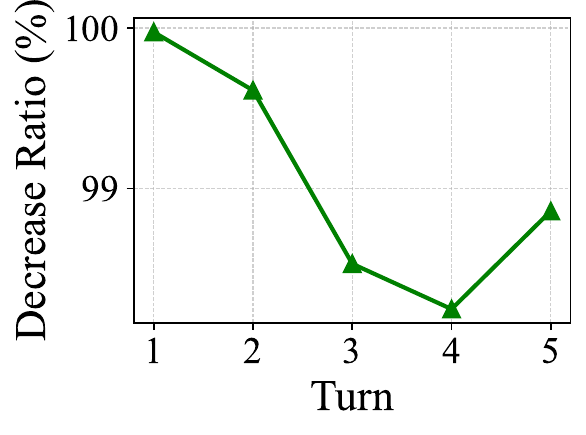}
        \caption{Fashion}
    \end{subfigure}
    \hfill
    \begin{subfigure}[t]{0.24\textwidth}
        \centering
        \includegraphics[width=\linewidth]{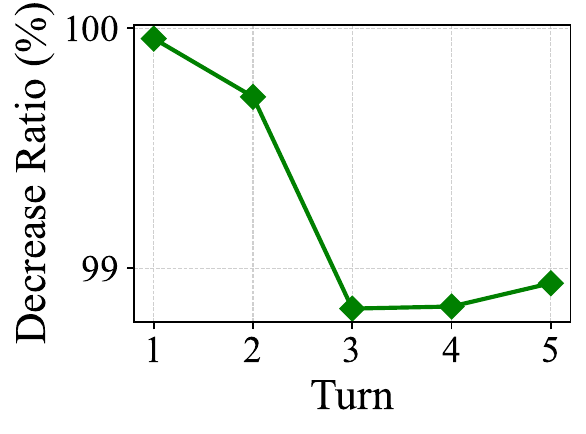}
        \caption{Home}
    \end{subfigure}
    \hfill
    \begin{subfigure}[t]{0.24\textwidth}
        \centering
        \includegraphics[width=\linewidth]{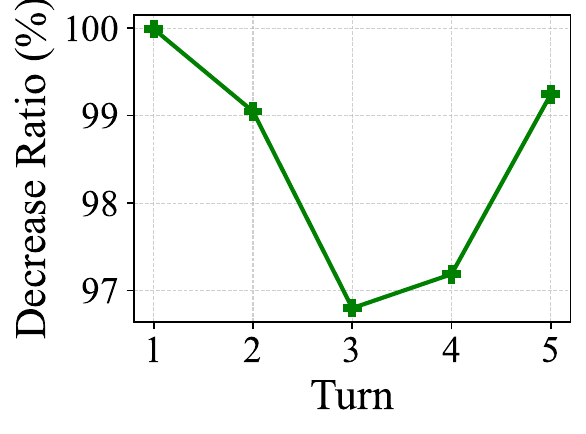}
        \caption{Inspired}
    \end{subfigure}
    \caption{Uncertainty Decrease Ratio.}
    \label{fig:uncertainty_reduction}
\end{figure}

To further understand how uncertainty-aware planning improves decision making, 
we analyze how quickly uncertainty over the candidate search space is reduced across interaction turns. Uncertainty is quantified as the entropy of the belief state, and we compare against a size-matched uniform distribution as a reference. The normalized decrease ratio is computed through:
\begin{equation}
\text{DecreaseRatio}_t = \frac{\mathcal{H}_{\text{ours},t}}{\mathcal{H}_{\text{uniform},t}},
\end{equation}
where $\mathcal{H}_{\text{ours}, t}$ denotes the uncertainty of the belief state induced by \method at turn $t$, and $\mathcal{H}_{\text{uniform},t}$ is the uncertainty of a uniform distribution over the same-sized candidate set, representing the uncertainty under random action selection.

Figure~\ref{fig:uncertainty_reduction} illustrates the decrease ratio across four datasets. \method reduces uncertainty substantially faster than the uniform baseline, with a sharp increase in the decrease ratio during early turns. This indicates that the belief is concentrated on a smaller set of candidates more quickly, enabling earlier and more confident decisions. This improvement is driven by uncertainty-guided planning. By selecting actions based on their expected long-term impact on uncertainty, \method eliminates irrelevant candidates more efficiently across turns, leading to higher success rates and fewer interaction steps.

Representative success and failure trajectories illustrating this behavior are shown in Appendix~\ref{sec:case_study}. We also discuss directions for extending \method in Appendix~\ref{sec:discussion}.

%%%%%%%%%%%%%%%%%%%%%%%%%%%%%%%%%%%%%%%%%%%%%%%%%%%%%%%%
\section{Conclusion}
\label{sec:conclusion}
%%%%%%%%%%%%%%%%%%%%%%%%%%%%%%%%%%%%%%%%%%%%%%%%%%%%%%%%
In this work, we study goal-oriented conversational decision making under uncertainty, emphasizing the need to coordinate information acquisition and target commitment over multiple turns. 
Existing approaches either rely on rigid structured formulations or flexible but myopic language models, limiting their ability to reason over long interaction trajectories. To address this, we formulate conversation as an uncertainty-aware sequential decision problem and propose \method, which integrates language models with structured planning by using uncertainty as a guiding signal. Experiments show that \method consistently improves both effectiveness and efficiency across datasets and model backbones. Further analysis reveals that uncertainty-aware planning enables faster belief concentration and more efficient information acquisition, highlighting its value for multi-turn decision making in conversational systems.

% %%%%%%%%%%%%%%%%%%%%%%%%%%%%%%%%%%%%%%%%%%%%%%%%%%%%%%%%
% \section*{Acknowledgments}
% %%%%%%%%%%%%%%%%%%%%%%%%%%%%%%%%%%%%%%%%%%%%%%%%%%%%%%%%
% Use unnumbered first level headings for the acknowledgments. All

% \section*{Ethics Statement}
% Authors can add an optional ethics statement to the paper. 
% For papers that touch on ethical issues, this section will be evaluated as part of the review process. The ethics statement should come at the end of the paper. It does not count toward the page limit, but should not be more than 1 page. 

\bibliography{colm2026_conference}
\bibliographystyle{colm2026_conference}

\clearpage
\appendix

\section{Case Study}
\label{sec:case_study}

\begin{figure}[h]
\centering
% \vspace{-5mm}
    \begin{subfigure}[t]{0.43\textwidth}
        \centering
        \includegraphics[width=\linewidth]{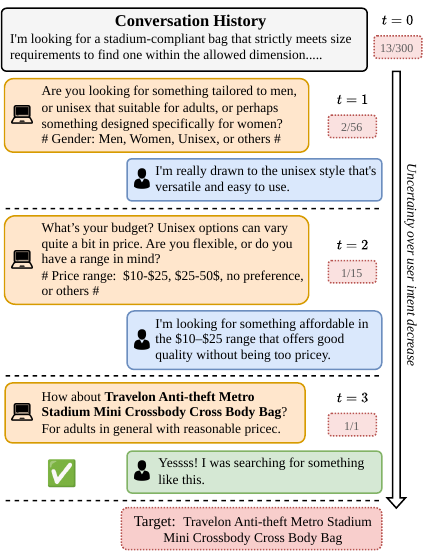}
        \caption{A Success Case from Fashion.}
        \label{fig:s_case}
    \end{subfigure}
    \hspace{0.05\textwidth}
    \begin{subfigure}[t]{0.43\textwidth}
        \centering
        \includegraphics[width=\linewidth]{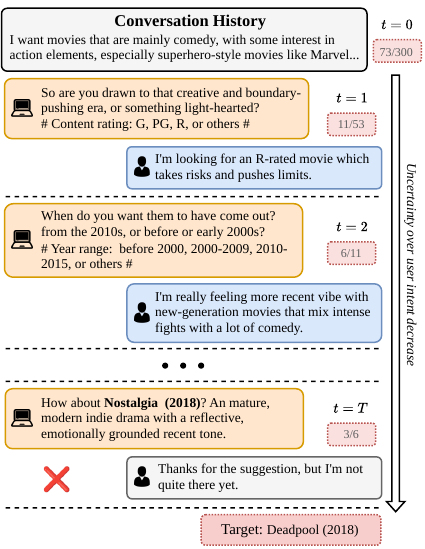}
        \caption{A Failure Case from Inspired.}
        \label{fig:f_case}
    \end{subfigure}
    \caption{Case study of \method framework.}
    \label{fig:case}
    % \vspace{-5mm}
\end{figure}

To better understand the mechanism of our \method framework, we illustrate two representative cases in Figure~\ref{fig:case}. 

In the success case (Figure~\ref{fig:s_case}), the agent effectively reduces uncertainty through clarification questions, rapidly narrowing the candidate search space and enabling a correct commitment within three turns. This demonstrates that uncertainty-aware action selection can guide efficient information acquisition and support accurate decision making.

In the failure case (Figure~\ref{fig:f_case}), although the target candidate is progressively promoted in the ranking (e.g., 73/300 → 11/53 → 6/11 → 3/6), the remaining ambiguity is not fully resolved. As a result, the agent fails to confidently identify the correct target and makes an incorrect commitment. This suggests that sometimes improving overall ranking alone is insufficient for reliable decision making, and points to a direction for future work in placing greater emphasis on distinguishing among top-ranked candidates, where residual uncertainty is most critical.
% We will explore more effective strategies to better resolve ambiguity among top candidates in the following work.

\section{Additional Training-Free Baselines}
\label{sec:additional_baselines}
We further compare \method against two training-free baselines that share only partial ingredients with our design. Prompt-MCTS~\citep{yu2023prompt} performs multi-step MCTS over LLM-proposed actions with LLM-sampled priors and an LLM-based outcome probe as the heuristic value, but does not maintain an explicit belief over candidates. INTENT-SIM~\citep{zhang2025clarify} decides when to clarify by sampling $N$ user intents and measuring entropy over them; we adapt it to our multi-turn setting by sampling $N{=}10$ intent responses per turn at temperature $0.5$, clustering by exact option match, and committing when the normalized entropy falls below a threshold. For both, the system LLM (Qwen3-4B), user simulator, turn budget ($T{=}5$), and candidate pool are kept identical to \method.

\begin{table}[h]
\centering
\caption{Head-to-head comparison against two additional training-free baselines on Qwen3-4B ($T{=}5$). Success rate (\%).}
\label{tbl:additional_baselines}
\footnotesize
\begin{tabular}{lcccc}
\toprule
Method & Beauty & Fashion & Home & Inspired \\
\midrule
Prompt-MCTS~\citep{yu2023prompt}     & 49.5 & 48.2 & 58.6 & 55.1 \\
INTENT-SIM~\citep{zhang2025clarify}  & 47.5 & 48.6 & 50.8 & 48.0 \\
\method (ours)                       & \textbf{82.1} & \textbf{83.4} & \textbf{84.1} & \textbf{89.8} \\
\bottomrule
\end{tabular}
\end{table}

Table~\ref{tbl:additional_baselines} shows that \method outperforms both baselines by a large margin. The gap is consistent with our ablation in Section~\ref{sec:ablation}: Prompt-MCTS has multi-step planning but no EIG signal (analogous to \method-NoEIG), while INTENT-SIM has neither multi-step planning nor structured belief tracking. \method's combination of both ingredients accounts for the observed gap.

% \section{More Ablation}
% \subsection{Ingredient}
% 

\section{Belief Initialization}
\label{sec:ablation_uniform_init}
To test whether \method's gains depend on the cosine-similarity warm start in Eq.~\ref{eq:belief_intial}, we re-run \method with a uniform initial belief $b_0(c_i) = 1/|\mathcal{C}_0|$ and leave the subsequent Bayesian updates unchanged.
\begin{table}[h]
\centering
\caption{Belief-initialization ablation on Qwen3-4B. Success rate (\%).}
\label{tbl:uniform_init}
\footnotesize
\begin{tabular}{lcccc}
\toprule
Initialization of $b_0$ & Beauty & Fashion & Home & Inspired \\
\midrule
Cosine similarity (default, Eq.~\ref{eq:belief_intial}) & \textbf{82.1} & \textbf{83.4} & \textbf{84.1} & 89.8 \\
Uniform                                                & 79.3 & 80.1 & 79.2 & 89.8 \\
\bottomrule
\end{tabular}
\end{table}

Under uniform initialization, \method's success rate drops by only a few points on Beauty, Fashion, and Home, while Inspired is unaffected (Table~\ref{tbl:uniform_init}). Even without an informative prior, \method still substantially outperforms the baselines in the main results. The drop concentrates on the three e-commerce domains, which have larger candidate pools and weaker attribute-item correlation, so removing the cosine warm start discards relatively more useful initial signal there; Inspired's smaller candidate space is largely insensitive to the choice of prior.

\section{Hyperparameter Sensitivity}
\label{sec:sensitivity}
To evaluate the robustness of \method's default configuration, we perform a one-at-a-time sensitivity sweep on each of the commitment thresholds $\theta$, $\epsilon$, and the MCTS search budget $K$, varying one parameter at a time around its validation-selected default.

\begin{table}[h]
\centering
\caption{One-at-a-time hyperparameter sensitivity sweep on Qwen3-4B. Success rate (\%).}
\label{tbl:sensitivity}
\footnotesize
\begin{tabular}{llcccc}
\toprule
Parameter & Value & Beauty & Fashion & Home & Inspired \\
\midrule
\multirow{4}{*}{$\theta$}   
& 0.6              & 82.8 & 83.7 & 86.0 & 91.8 \\
& 0.7              & 83.6 & 83.9 & 84.4 & 90.8 \\
& 0.8 (default)    & 82.1 & 83.4 & 84.1 & 89.8 \\
& 0.9              & 82.6 & 83.9 & 85.0 & 93.9 \\
\midrule
\multirow{3}{*}{$\epsilon$}
& 0.4              & 83.3 & 84.6 & 83.9 & 90.8 \\
& 0.5 (default)    & 82.1 & 83.4 & 84.1 & 89.8 \\
& 0.6              & 82.9 & 84.1 & 84.1 & 90.8 \\
\midrule
\multirow{4}{*}{$K$}
& 10               & 72.9 & 71.1 & 79.8 & 90.8 \\
& 25               & 80.7 & 80.4 & 83.9 & 86.7 \\
& 50 (default)     & 82.1 & 83.4 & 84.1 & 89.8 \\
& 100              & 84.3 & 84.0 & 85.5 & 92.9 \\
\bottomrule
\end{tabular}
\end{table}

Within the tested range, success rate varies by only a few points on each dataset across all four parameters (Table~\ref{tbl:sensitivity}). The defaults are selected on a validation set and remain competitive on the test sets without per-dataset tuning. For the MCTS budget $K$, success rate rises monotonically and largely saturates by $K=50$, which we adopt as a budget-quality trade-off.

\section{Computational Efficiency}
\label{sec:efficiency}
A key design choice of \method is that MCTS rollouts use a deterministic surrogate simulator (attribute lookup over candidates, no LLM calls), so the $K$ simulations per turn incur no LLM forward passes. The only LLM calls per turn are action proposal and utterance verbalization (plus an optional refined commit), a small fixed number that does not grow with the search budget $K$.

\begin{table}[h]
\centering
\caption{Computational efficiency of \method on Qwen3-4B using MCTS.}
\label{tbl:efficiency}
\footnotesize
\begin{tabular}{lccc}
\toprule
Dataset & $K$ & LLM calls per turn & MCTS time per turn (s) \\
\midrule
\multirow{3}{*}{Inspired}
& 10  & 2.62 & 4.00 \\
& 50  & 2.50 & 4.26 \\
& 100 & 2.47 & 4.57 \\
\midrule
\multirow{3}{*}{Home}
& 10  & 2.84 & 8.88 \\
& 50  & 2.71 & 9.39 \\
& 100 & 2.65 & 10.19 \\
\bottomrule
\end{tabular}
\end{table}

Table~\ref{tbl:efficiency} confirms this empirically. As $K$ grows from 10 to 100, the number of LLM calls per turn is low and slightly declines, since a larger search budget concentrates the belief faster and the planner triggers fewer refined-commit calls. MCTS wall-clock time per turn grows by only about 15\% over the same range.

\section{Discussion and Future Work}
\label{sec:discussion}
\paragraph{Extension to open-ended settings.}
Although we evaluate \method on tasks with a fixed candidate pool, the discretization is a property of the current problem setting, not of the \method formulation itself. \method only requires (i) a belief distribution over hypotheses and (ii) an EIG estimate per action; neither is intrinsically tied to a finite, enumerated candidate set. The framework extends to open-ended settings by (a) replacing the enumerated candidate set with a generative hypothesis space (e.g., LLM-sampled hypotheses following~\citep{piriyakulkij2024doing}) and (b) estimating EIG over sampled hypotheses via a Monte Carlo approximation of the same expression used in the discrete case (Eq.~\ref{eq:eig}), replacing the exact partition over candidates with sampled posteriors. Empirical exploration of this open-ended extension is an important direction for future work.

\paragraph{Extension to multimodal settings.}
Our current evaluation is text-only, but three of our four domains are e-commerce settings where product visuals carry information complementary to attribute text. Multimodal foundation models for recommendation~\citep{geng2023vip5}, goal-oriented multimodal interactive recommendation with verbal and non-verbal feedback~\citep{wu2023goal}, and recent e-commerce work on multimodal instruction data~\citep{ling2025captions} and strategic use of product visuals~\citep{ling2025ecommmmu} all point to the value of visual signals in this setting. Extending \method to incorporate such signals is a natural direction for richer e-commerce and product-search scenarios.

\paragraph{Refinements to commitment and initialization.}
Two internal components of \method admit natural generalizations. First, the entropy-based commitment trigger is itself a value-of-information (VoI) criterion in spirit, deciding to stop asking once the expected marginal information from another turn is small relative to the cost; richer VoI formulations that reason jointly about future EIG and downstream commitment reward, in the style of~\citep{li2024mediq, grand2026shoot}, are a natural direction for further generalization. Second, the cosine-similarity belief initialization provides only a modest warm start (Appendix~\ref{sec:ablation_uniform_init}); task-conditioned or learned initializations paired with matching planning schedules could allow the planner to benefit more from stronger priors when they are available.

\section{Prompts}
\label{sec:prompts_appendix}

We provide all prompts used by \method here. The same templates are used across all four datasets.

\subsection{User Simulator}
\label{sec:prompts_user}
The user simulator is an instruction-prompted Llama-3.2-3B-Instruct conditioned only on the target item. A leakage filter post-processes every generated response to rewrite any occurrence of the target name to a generic placeholder before returning the response to the agent.

\textbf{System prompt}
\vspace{4pt}
\hrule
\begin{lstlisting}
You are a {seeker_role} in a conversational {item_noun} recommendation system.
You are looking for a specific {item_noun} with these characteristics:
{attribute_bullets}

IMPORTANT: NEVER mention the exact name of your target {item_noun}.
Be conversational, concise (under 50 words).
\end{lstlisting}
\hrule

\textbf{Question response}
\vspace{4pt}
\hrule
\begin{lstlisting}
The recommender asks: {question_text}

Options:
A. {option_1}
B. {option_2}
C. {option_3}
D. {option_4}

Your preference is: {ground_truth_option}

Respond naturally. Do NOT mention the exact name of your target {item_noun}. Under 50 words.
\end{lstlisting}
\hrule

\textbf{Accept response}
\vspace{4pt}
\hrule
\begin{lstlisting}
The recommender suggests: {item}
This is exactly what you wanted! Accept enthusiastically. Under 30 words.
\end{lstlisting}
\hrule

\textbf{Reject response}
\vspace{4pt}
\hrule
\begin{lstlisting}
The recommender suggests: {item}
Not what you want. You want: {attribute_hints}.
Politely decline. Do NOT mention your target {item_noun}'s name. Under 50 words.
\end{lstlisting}
\hrule

\textbf{Leakage filter.} Post-generation, any case-insensitive occurrence of the target's name (and its name-variant prefixes, e.g., \texttt{"Joker (2019)" $\rightarrow$ "joker"}) is rewritten to the dataset-specific replacement (\texttt{[a product I like]} or \texttt{[a movie I like]}) before the response is returned to the agent.

\subsection{Agent Backbone}
\label{sec:prompts_agent}
The agent backbone LLM is invoked in three roles: action proposal (planning), utterance verbalization (execution), and refined commitment (when belief confidence is low at commit time).

% \vspace{7pt}
\textbf{System prompt}
\vspace{4pt}
\hrule
\begin{lstlisting}
You are a conversational recommendation assistant.
\end{lstlisting}
\hrule

% \vspace{7pt}
\textbf{Action proposal}
\vspace{4pt}
\hrule
\begin{lstlisting}
You are helping a conversational recommendation system decide what to ask the user.

Given the conversation so far and the remaining candidate {item_noun}s, propose options for EVERY available attribute below.

Conversation history:
{history}

Remaining candidates (sorted by belief probability):
{candidate_list}

Available attributes and their value distributions among candidates:
{attribute_summary}

Already asked: {asked_attributes}

For EACH attribute listed above, propose 2-4 options that would best help distinguish between candidates given the conversation context. Pick options that split the candidate set into roughly equal groups.

Respond in this exact format (one line per attribute):
ATTRIBUTE: <name> | OPTIONS: <option1>, <option2>, <option3>
ATTRIBUTE: <name> | OPTIONS: <option1>, <option2>, <option3>, <option4>
...

You MUST include ALL available attributes. Do NOT skip any.
\end{lstlisting}
\hrule

\textbf{Ask verbalization}
\vspace{4pt}
\hrule
\begin{lstlisting}
You are in a {context}.

Conversation so far:
{conversation_history}

You want to ask the user about their preference on {attribute_name}.
The options are:
A. {option_1}
B. {option_2}
C. {option_3}
D. {option_4}

Generate a natural, conversational question that asks about this attribute and presents these options. Be friendly and concise (1-2 sentences).
\end{lstlisting}
\hrule

\textbf{Commit verbalization}
\vspace{4pt}
\hrule
\begin{lstlisting}
You are in a {context}.

Conversation so far:
{conversation_history}

Top candidates considered: {candidate_name}

Based on the conversation, you believe {candidate_name} is the best match.
Generate a natural recommendation utterance presenting this {item_noun} to the user. Be confident and concise (1-2 sentences).
\end{lstlisting}
\hrule

\textbf{Refined commit}
\vspace{4pt}
\hrule
\begin{lstlisting}
You are in a {context}.

Conversation so far:
{conversation_history}

You must now commit to one {item_noun}. Here are the remaining candidates with their belief probabilities:
  - {candidate_1} (belief: {p_1})
  - {candidate_2} (belief: {p_2})
  ...

Think step by step:
1. What preferences has the user expressed so far?
2. Which candidate best matches those preferences?
3. Why is it the best match?

Format:
REASONING: <your step-by-step analysis>
COMMIT: <exact {item_noun} name>
\end{lstlisting}
\hrule

\end{document}